\documentclass[10pt,twocolumn,letterpaper]{article}
\usepackage[a4paper, margin=.8in, columnsep=.3in]{geometry}
\usepackage{standalone}
\usepackage{times}
\usepackage{epsfig}
\usepackage{graphicx}
\graphicspath{{figures/}}
\usepackage{amsmath}
\usepackage{amssymb}
\usepackage{color}
\usepackage{makecell}
\usepackage{comment}
\usepackage{booktabs}
\usepackage{mathtools}
\usepackage{tabu}
\usepackage{hhline}
\usepackage{xcolor}
\usepackage{colortbl}
\usepackage{url}

\def\abstract{\centerline{\large\bf Abstract}\vspace*{10pt}\it}

\usepackage[normalem]{ulem} 

\def\eg{\emph{e.g.}} 
\def\ie{\emph{i.e.}} 
\def\etal{\emph{et al.}}

\usepackage{multirow}
\usepackage{rotating}
\usepackage{protosem}

\newtheorem{rem}{Remark}

\newcommand{\bs}{\boldsymbol}

\newcommand{\cl}{\mathcal}
\newcommand{\ts}{\textstyle}

\DeclareMathOperator*{\argmin}{arg\,min}

\newcommand{\apc}[2]{%
	\ifmmode
	\mathrm{{\rm AP}^{#1}(#2)}%
	\else%
	AP$^{#1}(#2)$\@\xspace%
	\fi%
}

\newcommand{\gva}[1]{#1}
\newcommand{\SAGHc}[1]{#1}

\def\openpifpaf{OpenPifPaf~\cite{kreiss2021openpifpaf}}
\def\oracle{Oracle}
\def\DSL{DeepSportLab}

\usepackage{hyperref}

\usepackage{caption} 

\usepackage{fancyhdr} 
\fancypagestyle{firststyle}{
   \fancyhf{}
   \fancyhead[R]{\textsc{Paper ACCEPTED in British Machine Vision Conference 2021's proceedings.}}
}

\usepackage{authblk}

\title{DeepSportLab: a Unified Framework for Ball Detection, Player Instance Segmentation and Pose Estimation in Team Sports Scenes}
\author[1]{Seyed Abolfazl Ghasemzadeh}
\author[1,2]{Gabriel Van Zandycke}
\author[1,2]{Maxime Istasse}
\author[1]{Niels Sayez}
\author[1]{Amirafshar Moshtaghpour}
\author[1]{Christophe De Vleeschouwer}
\affil[1]{UCLouvain ICTEAM/ELEN Belgium \authorcr {\small\nolinkurl{<firstname>.<lastname>@uclouvain.be}}}
\affil[2]{SportRadar AG \authorcr {\small\nolinkurl{<f>.<lastname>@sportradar.com}}}
\date{}

\begin{document}
\maketitle\thispagestyle{firststyle}

\begin{abstract}
	This paper presents a unified framework to \textit{(i)} locate the ball, \textit{(ii)} predict the pose, and \textit{(iii)} segment the instance mask of players in team sports scenes. Those problems are of high interest in automated sports analytics, production, and broadcast. A common practice is to individually solve each problem by exploiting universal state-of-the-art models, \eg, Panoptic-DeepLab for player segmentation. In addition to the increased complexity resulting from the multiplication of single-task models, the use of the off-the-shelf models also impedes the performance due to the complexity and specificity of the team sports scenes, such as strong occlusion and motion blur. To circumvent those limitations, our paper 
	proposes to train a single model that simultaneously predicts the ball and the player mask and pose by combining the part intensity fields and the spatial embeddings principles. Part intensity fields provide the ball and player location, as well as player joints location. Spatial embeddings are then exploited to associate player instance pixels to their respective player center, but also to group player joints into skeletons.  
	We demonstrate the effectiveness of the proposed model on the DeepSport basketball dataset, achieving comparable performance to the SoA models addressing each individual task separately.
\end{abstract}

\section{Introduction}

The automation of team sports analytics and broadcasting~\cite{chen2010, chen2011, fernandez2010}, relies on detailed scene interpretation, which itself depends on the ability to detect the ball, segment the players (e.g. for improved tracking and recognition), and predict their pose (e.g. for action recognition).
Our work leverages Convolutional Neural Networks (CNNs) to tackle those tasks. 

The most natural approach to solve the three tasks above with CNNs is to use a pre-trained state-of-the-art model for each individual problem, \eg, Mask R-CNN~\cite{he2017maskrcnn}, PifPaf~\cite{kreiss2019pifpaf}, and Panoptic-DeepLab~\cite{cheng2020panoptic} for ball localization, player pose estimation, and player instance segmentation, respectively.
Such approach, however, results in poor performance since team sports scenes -- especially those of indoor sports -- are more complex compared with in-the-wild scenes.
First, they involve strong player occlusions, \eg, a defending player blocking an attacking one. Such occlusions are usually considered as ``crowd'' in in-the-wild datasets, such as CityScapes; hence, excluded from training.
Second, they contain fast moving players and balls causing motion blur.
Third, dealing with sports players is subject to deformation, since the players often jump, run, or stretch their bodies.
Fourth, due to its frequent interactions with players, the ball is often partially occluded.
Fifth, indoor team sports scenes suffer from weak contrast between the ball and field and from the reflection of the players in the field.

An improved approach consists in fine-tuning the weights of pre-trained models with task-specific datasets~\cite{van2019real}. 
Nevertheless, that solution has its shortcomings. General purpose models are often demanding in terms of the memory and computation, which prevents their real-time application, especially if multiple models have to run in parallel. Moreover, by tackling each task individually, the CNN model ignores the correlation between them, which might hamper their performance~\cite{kendall2018multi}.

\begin{figure*}[t]
	\centering
	\includegraphics[width=\textwidth]{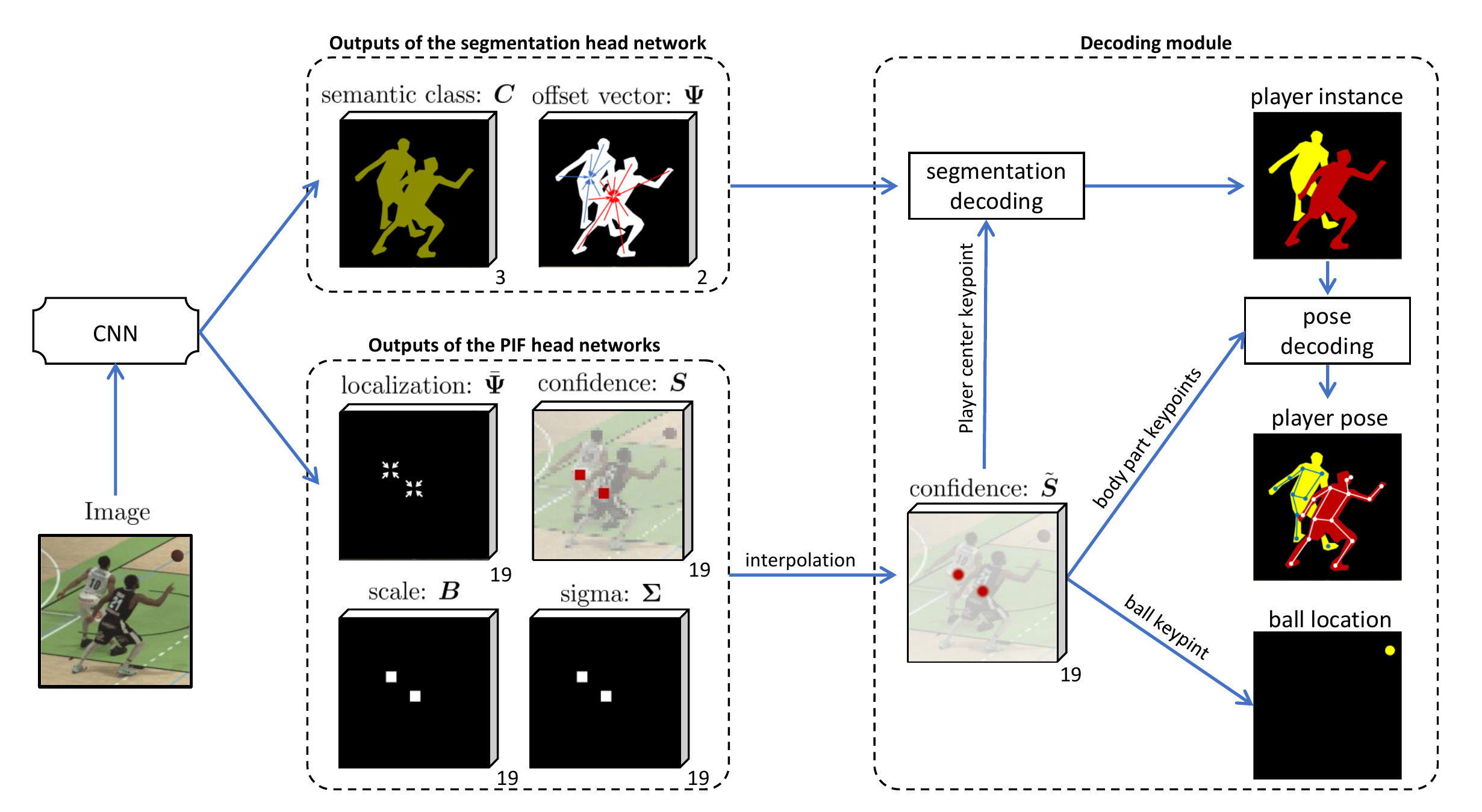}
	\caption{\textbf{An overview of DeepSportLab.} Our CNN outputs: \textit{(i)} semantic class scores, \textit{(ii)} offset vectors to associate pixels to their instance center, and \textit{(iii)} a set of low resolution outputs used to generate the high-resolution keypoint confidence maps. This set includes the keypoints confidence map, the keypoints localization vectors, the size (or width) of the keypoints, and the scale parameters, which is used for scaling the per-keypoint localization loss (see Eq.~\eqref{eq:loss_offset}). The segmentation decoding module fuses the first two outputs and the high-resolution confidence map (\ie, only the map of the center-of-the-player keypoint) in order to predict the player masks. The confidence maps of the body part keypoints and the resulting player masks are then fused by the pose decoding module and yield the player poses. Finally, the location of the ball is extracted from the high-resolution map of the ball keypoint. Such pipeline results in a light-weight decoding process. Note that the images in this figure are only for visualization purposes and are neither the ground truth annotations nor the predicted maps. In the outputs of the PIF head networks, we only show the maps associated with the center-of-the-player keypoint.}
	\label{fig:model}
\end{figure*}

In this work, we propose to rely on a single CNN  to jointly localize the ball and predict the player poses and player instance masks, given a single input image. An overview of our model is presented in Fig.~\ref{fig:model}. As an important feature of the proposed model, we define an extended set of 19 keypoint-types including 17 human body parts, ball centroid, and player centroid; hence, treating the player centroids and ball detection tasks as keypoint detection problems. Our CNN is illustrated in more details in Fig.~\ref{fig:cnn}, where it outputs two sets of predictions. Inspired by Panoptic-DeepLab~\cite{cheng2020panoptic}, one head network predicts pixel-wise semantic classes and offset vectors, providing the information required to associate player pixels to their centroid. As shown in Fig.~\ref{fig:model}, the other three head networks predict the Part Intensity Field (PIF) of the PifPaf framework~\cite{kreiss2019pifpaf} for the 19 keypoint-types, \ie, a collection of confidence score, localization vectors, and the size and scale of that keypoint. Our instance segmentation decoder fuses the semantic classes, offset vectors, and player centroids. The resulting player mask is then leveraged to associate the predicted body parts to the individual players. Therefore, unlike~\cite{kreiss2019pifpaf}, our specialized model does not require the Part Affinity Field (PAF) for building the player skeletons; resulting in a light-weight decoding procedure.
We train our model on the combination of the COCO~\cite{lin2014microsoft} and DeepSport~\cite{deepsport} datasets. The software is made freely available at \url{https://github.com/ispgroupucl/DeepSportLab}.

\section{Related works}
\label{sec:related-works}

\paragraph{Instance segmentation.} Instance segmentation systems can be grouped into  \textit{proposal-based}~\cite{girshick2014rcnn,girshick2015fastrcnn,he2017maskrcnn} and \textit{proposal-free}~\cite{newell2017associative,cheng2020panoptic} methods. The common idea in the former type of approaches is to predict a number of object proposals (or bounding-boxes) and then to classify the objects within each of those proposals. In contrast, most of the works in the latter category (including the proposed DeepSportLab) rely on  predicting  pixel-wise  embedding vectors such that pixels belonging to an instance receive similar embeddings. In that context, a clustering algorithm is exploited to group the pixels based on the similarity between their embedding vectors, and in turn, to create the instance masks. The first proposal-free method is reported in~\cite{liang2017proposal}, where the CNN outputs the pixel-wise coordinates of the top-left and bottom-right corners of the corresponding object bounding-box. Those coordinates can be thought as 4-D embeddings. Newell \etal~\cite{newell2017associative} introduced associative embeddings, which can be thought as a vector representing each pixel's cluster. In particular, the embedding vectors in~\cite{newell2017associative} are related to an (non-physical) abstract space and are supervised without ground truth references -- unlike the formalism in~\cite{liang2017proposal}. Other intermediate approaches have been proposed in~\cite{fathi2017semantic,kong2018recurrent,brabandere2017semantic} with different clustering algorithms and loss functions. Novotny
\etal~\cite{novotny2018semi} extended the concept of associative embedding to spatial embedding: the network predicts pixel-wise offset vectors such that pixels of an instance point to that instance's center.
That idea later inspired several works~\cite{kendall2018multi,uhrig2018box2box,neven2019instance}, specially Panoptic-DeepLab~\cite{cheng2020panoptic}, which is one of the current top-performing instance segmentation models. Inspired by~\cite{cheng2020panoptic}, our network predicts offset vectors from each pixel to its corresponding player instance center, which will be used to form the player instances.

\paragraph{Human pose estimation.}
Human pose estimation approaches can be divided into two groups: \textit{top-down}~\cite{papandreou2017towards,he2017maskrcnn} and \textit{bottom-up}~\cite{kreiss2019pifpaf,kocabas2018multiposenet,papandreou2018personlab,newell2017associative,cao2019openpose}. Top-down methods first perform a human detector and then locate the body parts within every detected bounding-box. Bottom-up methods (\ie, the context of this paper), on the contrary, first estimate body parts followed by forming the skeleton.

OpenPose~\cite{cao2019openpose,cao2017realtime} model revolutionized the bottom-up approaches by showing that a non-parametric representation (\ie, PAF) can be learned to associate body parts with individual human instances. Newell \etal~\cite{newell2017associative} used the notion of associative embedding, which can be thought as a vector representing each keypoint's cluster. In particular, keypoints with similar embedding vectors are assigned to the same skeleton. 
PersonLab~\cite{papandreou2018personlab} integrates the human pose estimation with instance segmentation. It learns to predict relative displacement of body parts, \ie reminiscent of the idea of spatial embedding~\cite{novotny2018semi}, allowing to group body parts into human skeleton. By combining and extending the use of short-range offset vectors (as in PersonLab) and PAF (as in OpenPose), Kreiss \etal \cite{kreiss2019pifpaf} presented PifPaf framework, which reaches excellent performance for crowded scenes. Our work adopts the PIF part of that framework for predicting the location of the keypoints. Regarding the association of the body part keypoints to the poses, we follow the Panoptic-DeepLab's formalism. Unlike PersonLab, where human instance masks are computed from the human poses (in combination with predicted long-range offset vectors and semantic segmentation), in this work, we leverage the predicted player instance masks to group the keypoints into player poses.

\paragraph{Deep learning applications in sports.} Deep learning has recently offered promising solutions for sports production, such as jersey number recognition~\cite{gerke2017soccer}, segmentation of the field, players, and lines~\cite{cioppa2018bottom}, player discrimination~\cite{lu2018lightweight,istasseassociative2019}, event segmentation~\cite{einfalt2019frame}, player detection~\cite{cioppa2020multimodal,csah2018evaluation,pobar2018mask}, swimming stroke rate detection~\cite{victor2017continuous}, player pose estimation~\cite{zecha2019refining,zecha2018kinematic,hwang2017athlete,fastovets2013athlete}, and ball localization~\cite{van2019real,reno2018convolutional,speck2016ball}. In particular, the classifier in~\cite{reno2018convolutional} decides whether a patch of an image contains a tennis ball. The authors in~\cite{speck2016ball} and~\cite{van2019real} propose to predict the position of the ball by formulating the problem as, respectively, a regression and segmentation task. In this paper, however, we treat ball localization as a keypoint detection problem. Unlike those works on player pose estimation, which are designed for scenes containing a single swimmer~\cite{zecha2019refining,zecha2018kinematic} or athlete~\cite{hwang2017athlete,fastovets2013athlete}, our proposed method aims at multi-player pose estimation in team sports scenes. 

\section{Proposed Method: DeepSportLab}\label{sec:method}

\paragraph{Notations.}Domain dimensions are represented by capital letters, \eg $P$. Tensors are denoted by  upper case bold symbols. 
Ground truth data or associated parameters are distinguished by an asterisk, \eg, $\bs N^*_{\rm ply}$ and $\bs N_{\rm ply}$ are, respectively, the number of annotated and predicted players. The set of keypoint-types is denoted by $\cl K \coloneqq \cl K_{\rm part} \cup \{{\rm ball}, {\rm ply}\}$, where ``${\rm ply}$'' denotes the center-of-the-player keypoint and $\cl K_{\rm part}\coloneqq \{{\rm left~eye}, \cdots, {\rm right~ankle}\}$ is the set of 17 body parts. 

\subsection{Principle} \label{sec:method-principle}
Given an input image with $P$ pixels and $N_{\rm ply}$ players, the goal of DeepSportLab is to predict \textit{(i)}~the location  of the ball; \textit{(ii)}~the instance mask \gva{(or set of pixels)} of each player $\cl I_i \subset \{1,\cdots,P\}$ for $i\in \{1,\cdots,N_{\rm ply}\}$; and \textit{(iii)}~the skeleton of players. 

As illustrated in Fig.~\ref{fig:model}, our CNN (detailed in Sec.~\ref{sec:method-architecture}) outputs two groups of parameters. One group consists of the predictions from segmentation head network: semantic class scores $\bs C \in [0,1]^{P}$ (\ie player vs. non-player) and offset (or spatial embedding) vectors $\bs \Psi\in \mathbb R^{P\times 2}$, \ie, displacement of each pixel from the center of a player it belongs to (similarly to Panoptic-DeepLab~\cite{cheng2020panoptic}). The second group of outputs includes PIF predictions for each keypoint-type $k\in \cl K$, \ie, a (low-resolution) pixel-wise confidence map $\bs S(k) \in [0,1]^{\bar{P}}$, a vector component (or localization vector) $\bar{\bs \Psi}(k) \in \mathbb{R}^{\bar{P}\times 2}$ pointing to the closest keypoint, size of the keypoint $\bs \Sigma(k) \in \mathbb{R}^{\bar{P}}$, and a scale parameter $\bs B(k) \in \mathbb{R}^{\bar{P}}$ (we set $\bar{P} = P/64$ in our experiments). As it will be detailed in Sec.~\ref{sec:method-inference}, high-resolution confidence maps are then generated by fusing the low-resolution confidence maps with the localization vectors and keypoint sizes. Such formulation results in a light network architecture, since the number of parameters for each keypoint-type at the network output is 
$5\bar{P} = 0.078P$. Moreover, the parameter $\bs B(k)$ is used as weights in the training loss to balance the localization error with respect to each keypoint-type (see below).

The segmentation decoding module then predicts the instance masks of the players by fusing the semantic class scores, offset vectors, and confidence maps of the center-of-the-player keypoint. Furthermore, the pose decoding module leverages the player instance masks to group the predicted body parts into individual skeletons. Finally, the location of the ball is extracted from the confidence map of the ball keypoint. 
\begin{figure*}[t]
	\centering
	\includegraphics[width=\textwidth]{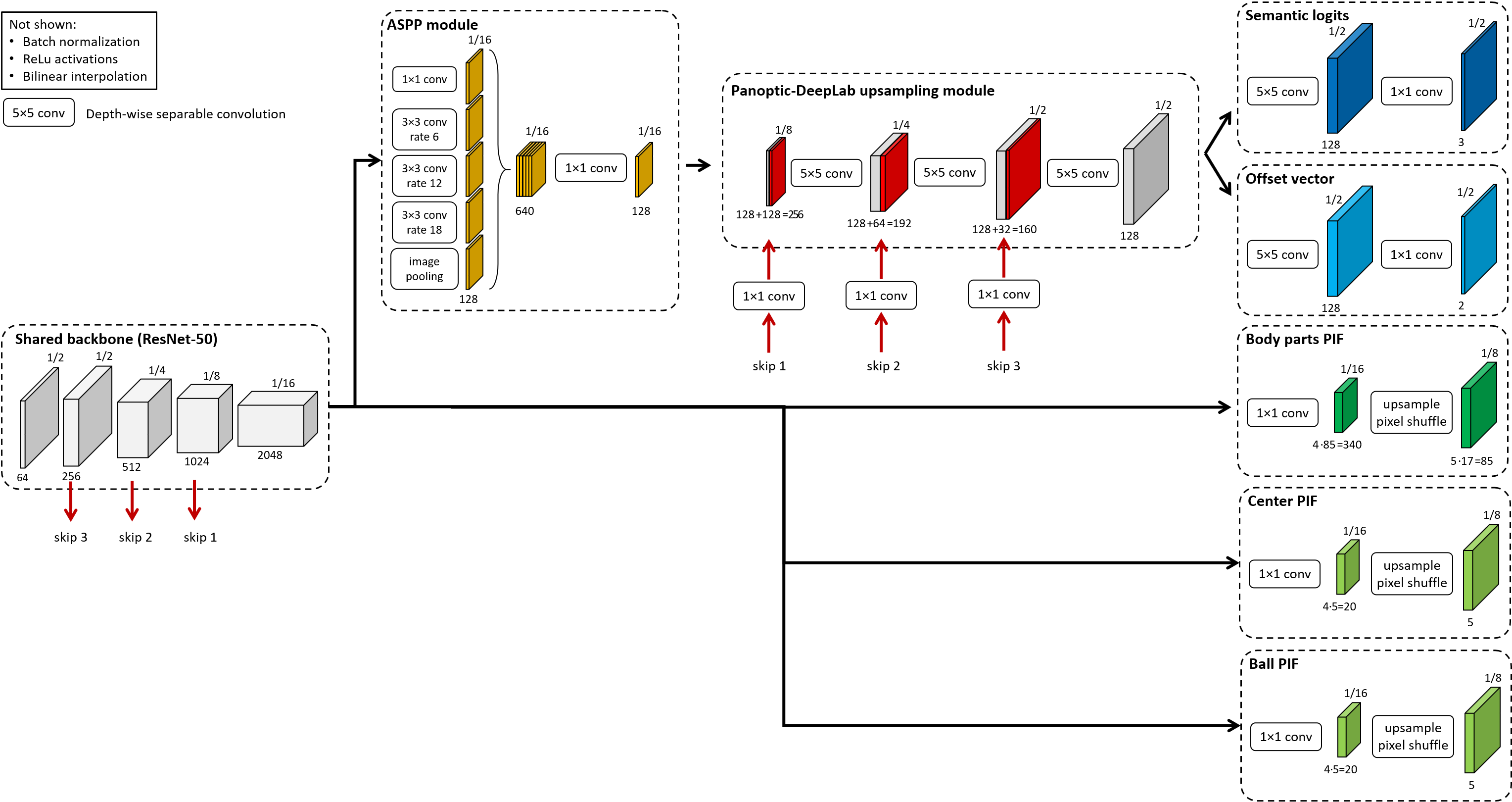}
	\caption{\textbf{CNN architecture of DeepSportLab.} Our network adopts an ASPP module, an upsampling module, and a dual-branch for the segmentation head. The upsampling module involves a bilinear interpolation before each concatenation step. The network also adopts a dual-PIF head network outputting the PIFs for the 19 keypoint-types. For lighter visualization, we concatenated the three PIF outputs in Fig.~\ref{fig:model}.}
	\label{fig:cnn}
\end{figure*}

\subsection{Network architecture}
\label{sec:method-architecture}

As illustrated in Fig.~\ref{fig:cnn}, the CNN module of DeepSportLab has one shared backbone, \ie, a ResNet-50 network~\cite{he2016deep}, for its four head networks. The segmentation head network consists of an Atrous Spatial Pyramid Pooling (ASPP) module involving Atrous convolutions~\cite{yu2015multi} to extract denser feature maps, a modified version of the upsampling module from Panoptic-DeepLab, and two task-specific prediction branches, which outputs the semantic logits and offset vectors, respectively. The upsampling module gradually increases the resolution of the features by leveraging the three skip connections from the backbone. In that module, a bilinear interpolation is applied before each concatenation step. Inspired by the PifPaf framework~\cite{kreiss2019pifpaf}, our three PIF head networks output the PIF parameters for, respectively, the ball, the center of the players, and the other 17 keypoint-types. Each PIF head is equipped with a pixel shuffle~\cite{shi2016real} operation upsampling the feature maps by a factor of two. The semantic logits and the offset vectors are further upsampled by a bilinear interpolation to reach the input image resolution before being fed to the decoding module.

Note that in Fig.~\ref{fig:model} the outputs of the three PIF head networks are concatenated for the sake of simpler visualization. Similarly, the PIF tensors, \eg, $\bs S(k), \bs \Psi(k)$ are defined for all 19 keypoint-types $k$. We recall that our segmentation and PIF head networks are adapted from the Panoptic-DeepLab~\cite{cheng2020panoptic} and PifPaf~\cite{kreiss2019pifpaf} frameworks, respectively. 

\subsection{Network supervision}\label{sec:method-training}
DeepSportLab requires a collection of ground truth data for its supervised training, \ie, binary semantic class scores $\bs C^* \in \{0,1\}^{P}$, player and ball instance masks, pixel-wise offsets to players' instance centroids $\bs \Psi^*\in \mathbb R^{P\times 2}$, and per-keypoint-type (low-resolution) confidence maps $\bs S^*(k)\in \{0,1\}^{\bar{P}}$, localization vectors $\bar{\bs \Psi}^*(k)\in \mathbb R^{\bar{P}\times 2}$, and keypoint sizes (or sigma) $\bs \Sigma^*(k) \in \mathbb{R}^{\bar{P}}$. Note that $\bs S^*(k), \bar{\bs \Psi}^*(k)$, and $\bs \Sigma^*(k)$ take non-zero values on only 16 neighboring pixels of each keypoint. The ground truth size parameters are the body part standard deviation values (provided by the COCO dataset) scaled according to the size of the keypoints in the image. 

Given the above-mentioned ground truth data, DeepSportLab is trained by minimizing the following loss function:
\begin{equation}\begin{split}
\cl L = & w_{\rm sem} \cl L_{\rm sem} + w_{\rm off} \cl L_{\rm off}\\ 
+ & w_{\rm cnf} \cl L_{\rm cnf} + w_{\rm loc} \cl L_{\rm loc} + w_{\rm sig} \cl L_{\rm sig}, \label{eq:loss}
\end{split}
\end{equation}
where the first two (and the last three) losses correspond to the segmentation (resp., pose) head network. In~\eqref{eq:loss}, $w_{\rm sem}$, $w_{\rm off}$, $w_{\rm cnf}$, $w_{\rm loc}$, $w_{\rm sig}$, are the semantic, offset, confidence, localization, and sigma loss weights, respectively, and
\begin{align}
\ts \cl L_{\rm sem} &= \frac{1}{P}\sum_{p}{\rm BCE}(\bs C^*(p), \bs C(p)),\nonumber\\
\ts \cl L_{\rm off} &= \frac{1}{P}\sum_{p}\|\bs \Psi(p)-\bs \Psi^*(p)\|_2^2,\nonumber\\
\ts \cl L_{\rm cnf} &= \sum_{p,k}{\rm BCE}(\bs S^*(k;p),\bs S(k;p)),\nonumber\\
\ts \cl L_{\rm sig} &= \sum_{p,k}|\bs \Sigma(k;p)-\bs \Sigma^*(k;p)|,\nonumber\\ 
\ts \cl L_{\rm loc} &= \sum_{p,k}\frac{|\bar{\bs \Psi}(k;p)-\bar{\bs \Psi}^*(k;p)|^2}{\bs B^2(k;p)} + \log(\bs B(k;p)), \label{eq:loss_offset}
\end{align}
where $\rm BCE$ denotes the binary cross entropy functions. We consider the mean squared error and mean absolute error for the rest of the sub-losses, summed over the defined variables, \eg, over the keypoints present in the image. For the localization loss we adapt the learnable standard deviation for the regression loss, \ie, inspired by \cite{kendall2018multi}. The purpose of such formalism is to let the network balance the localization error based on the size of the keypoint. Intuitively, while a localization error might be minor for a large keypoint, it can be major for a small keypoint. 

In practice, for the sake of numerical stability, we train the network to predict the logarithm of $\bs B$ and $\bs \Sigma$\gva{, and use the exponential of the prediction}. By doing so, both the localization loss $\cl L_{\rm loc}$ and the fusion operator (see Eq.~\eqref{eq:conf} below) avoid any division by zero during the training and inference, respectively.

\subsection{Inference} \label{sec:method-inference}
DeepSportLab consists in two main inference sub-tasks: \textit{(i)} keypoint decoding, which provides the center of the ball and players as well as the body parts keypoints; \textit{(ii)} player instance segmentation, which also enables pose recognition, as it assigns an instance label to each of the body part keypoints.

Recall that the keypoint confidence maps $\bs S(k)$ are coarse. In order to obtain high-resolution maps, they are fused with the localization vectors $\bar{\bs \Psi}(k)$ and sigma parameters $\bs \Sigma(k)$ through a convolutional operator with unnormalized Gaussian kernel, \ie,
\begin{equation}\label{eq:conf}
\tilde{\bs S}(k;p) \coloneqq \ts\sum_{u} \bs S(k;u)~\exp{\big(-\ts\frac{\|p-(u+\bar{\bs \Psi}(k;u))\|^2}{2\bs \Sigma^2(k;u)}\big)},
\end{equation} 
where $p$ and $u$ are the pixel coordinates in, respectively, high- and low-dimensional pixel grids. The location of the ball is identified by finding the maximum value in the confidence map of the ball keypoint. This amounts to a top-1 detection strategy.

For the next decoding task, each pixel $p$ classified as a player is assigned to only one player instance using a simple center regression. Equivalently, pixels are grouped into individual instances according to their displacements from the center of each player. Concretely, the mask of each player $i$ writes
\begin{equation}\begin{split}
\cl I^{\rm ply}_i  \coloneqq \biggl\{ p : & \tilde{\bs C}(p) \equiv {\rm player}, \\
& i = \ts\argmin_{j}\left\| o^{\rm ply}_j - (p+\bs \Psi(p))\right\|_2 \biggl\} . \label{eq:baseline_grouping}
\end{split}
\end{equation}
For each of the other 18 keypoint-types $k \in \cl K\backslash \{\rm ball\}$, we then compute a set of keypoint instances $\cl O^k \coloneqq \{o_1^k,\cdots, o_{N_k}^k\}$, with $N_k$ denoting the number of detected instances for keypoint-type $k$, by finding the maxima within each player's mask in the high-resolution confidence map $\tilde{\bs S}(k)$. We discard the detected keypoint instances with the confidence score less than 0.1.
Having identified the mask of each player, the player pose can be simply decoded: the skeleton of a player is formed by collecting body part keypoints whose coordinates belong to the mask of that player. 

If multiple keypoint instances of the same type are assigned to one skeleton, we keep only the keypoint instance with the highest confidence score.

\begin{rem}
Notice that our pose decoding algorithm described above is very fast and simple. Compared with PifPaf~\cite{kreiss2019pifpaf}, DeepSportLab does not require to learn the part affinity fields with dimension $\bar{P}\times 19 \times 7$ for pose decoding. Moreover, the greedy decoding algorithm in PifPaf~\cite{kreiss2019pifpaf}, which is similar to the  PersonLab's~\cite{papandreou2018personlab}, starts from a body part and finds the next connected body part (among other candidate parts, given the part affinity fields) by computing a pixel-wise association score. 
\end{rem}


\begin{table*}[h!]
\centering
\scalebox{0.75}{
\begin{tabular}{|l|c|cc|ccc|cc|}
	\hline
	\multirow{2}{*}{Method}
	&
	\multirow{2}{*}{$ {\rm bDQ}$}
	&
	\multirow{2}{*}{$ {\rm pSQ}$}
	&
	\multirow{2}{*}{$ {\rm pDQ}$}
	&
	\multicolumn{3}{c|}{$ {\rm pEQ}$}
	&
	\multirow{2}{*}{$ {\rm ms}$}
	&
	\multirow{2}{*}{$ {\rm MB}$}
	\\
	\hhline{*{2}{|~|}*{2}{~}*{3}{|-}}
	& &	& &  $ {\rm AP}$ & $ {\rm AR}$ & $ {\rm F}_1$ && \\
	\Xhline{2pt}
	\rowcolor{blue!10}	DeepSportLab & \gva{52.07} & 80.3 & 90.1 & 87.5 & 82.1 & 42.4 & \SAGHc{$436 \pm 105$} & \SAGHc{1757} \\
	\hline
	Pan.-DeepLab~\cite{cheng2020panoptic} & -- & 82.2 & 91.4 & -- & -- & -- & \SAGHc{$69 \pm 7$} & \SAGHc{1809} \\\hline
	\openpifpaf & -- & -- & -- & 88.5 & 79.6 & 41.9 & \SAGHc{$155 \pm 18$} & \SAGHc{1623} \\\hline
	BallSeg~\cite{van2019real} & \gva{46.16} & -- & -- & -- & -- & -- & \gva{$14 \pm 6$} & \gva{1239} \\
	\Xhline{2pt}
\end{tabular}
}
\caption{\textbf{Performance on DeepSport's \textit{test}:} Comparison of the proposed multi-task DeepSportLab with SoA models addressing each individual tasks. \gva{Panoptic-Deeplab and BallSeg are trained on the DeepSport dataset, while PifPaf is trained on COCO. DeepSportLab uses both DeepSport and COCO datasets to train the player and ball instances segmentation and only uses COCO to train the player pose.}
\gva{DeepSportLab compares favorably against its counterparts in term of quality measures, while reducing by a factor 3 the required memory.}}
\label{tab:perf}
\end{table*}

\begin{table*}[h]
\centering
\scalebox{0.75}{
\begin{tabular}{|cc|c|cc|ccc|}
	\hline
	\small player & \small player & \multirow{2}{*}{Decoder} & \multirow{2}{*}{pSQ}&
	\multirow{2}{*}{pDQ} & \multicolumn{3}{c|}{pEQ}
	\\\hhline{|~~|~|~~|---|}
	\small keypoints & \small masks & & & &$ {\rm AP}$ & $ {\rm AR}$ & $ {\rm F}_1$ \\
	\Xhline{2pt}\rowcolor{blue!10}
	 \DSL&\DSL & \multirow{3}{*}{DeepSportLab} & 80.3 & 90.1 & 87.5 & 82.1 & 42.4 \\
	 \hhline{--~-----}
	 \DSL & \oracle  && 100  & 100  & 87.7  & 83.8 & 42.9 \\
	 \hhline{--~-----}
	 \openpifpaf & Pan.DeepLab~\cite{cheng2020panoptic} && \SAGHc{82.2} & \SAGHc{91.4} & \SAGHc{87.3} & \SAGHc{82.1} & \SAGHc{42.3}
	 \\\Xhline{2pt}
\end{tabular}
}
\caption{\textbf{DeepSportLab Decoder study.} Oracle data or off-the-shelf models are used to analyse the sensitivity of DeepSportLab's decoder to the masks and keypoints predictions. The use of instance ground-truth masks show that the DSL instance segmentation is good enough to decode the player pose. DSL also compares favorably to a decoding using a combination of \openpifpaf\ and Pan.DeepLab~\cite{cheng2020panoptic}.}
\label{tab:decoder}
\end{table*}

\section{Experiments}\label{sec:experiments}
Our model is trained using a mix of the COCO~\cite{lin2014microsoft} and DeepSport~\cite{deepsport} datasets and is compared against Panoptic DeepLab~\cite{cheng2020panoptic} for player instance segmentation, \openpifpaf\ for player pose estimation, and BallSeg~\cite{van2019real} for ball detection.

\paragraph{DeepSport dataset.} It consists of 672 images of professional basketball games captured from 29 arenas involving a large variety of game configurations and various lighting conditions \cite{deepsport}. Each image captures half of the basketball court with a resolution between 2Mpx and 5Mpx. The resulting images have a definition varying between 65px/m (furthest point on court in the arena with the lowest resolution cameras) and 265px/m (closest point on court in the arena with the highest resolution cameras).
We extract 100 (out of 672) images for the validation and 64 images for the testing such that the arenas in the test set are neither present in the training nor the validation sets. 
The dataset contains approximately 380 annotations of ball masks and 5500 annotations of player masks and poses. The poses (composed of only 4 keypoint-types) are only used for evaluation.
The images are scaled to keep humans with similar height compared to those found in the COCO dataset~\cite{lin2014microsoft}.

\paragraph{COCO dataset.} In order to learn the pose of players, we consider a subset of the COCO dataset~\cite{lin2014microsoft} consisting of images containing only humans or balls. We also filter out the images containing at least one person with the area of larger than 10\% of the whole image. This filtering results in 42271 and 2356 images for the training and validation, respectively.

\paragraph{Mismatch between DeepSport and COCO pose annotations.} The COCO dataset, used to train the body part keypoints, contains pose annotations with 17 body parts; whereas, in DeepSport dataset used at testing, they are identified by four body parts: head, hip, foot~1, and foot~2, \ie, agnostic about the skeletons facing forward or backward. 
Hence, the quality computation phase during testing requires a delicate treatment. First, the skeletons predicted with the COCO convention are converted into the DeepSport skeleton format as follows. The locations of the head and hip for metric computations are defined, respectively, as the middle of the left- and right-ear and the middle of the left- and right-hip. Since foot~1 and foot~2 labels are interchangeable in DeepSport dataset,
\gva{the keypoint metric (sensitive to inversion)}
is computed for the two assignments of the feet, \ie, assigning foot~1/foot~2 to either left-ankle/right-ankle or right-ankle/left-ankle, and the assignment with a higher resulting \gva{value} is considered. We note that this conversion of skeleton introduces error in quality computation of predicted poses, but since that computation is the same for different models, performed comparisons remain fair.

\paragraph{Training setup.} Each batch of data during training involves equal share of images from DeepSport and COCO datasets. Therefore, during every epoch, the learning algorithm works over a different subset of the COCO dataset.
We trained our models with SGD optimizer with a learning rate set to $3\cdot 10^{-4}$, momentum to $0.95$, and weight decay to $10^{-5}$. We perform random horizontal flip, scaling, and rotation followed by a random $641\times 641$ cropping during training with batch size of $4$. We set the weights of the training sub-losses as $w_{\rm sem} = 10$, $w_{\rm off} = 0.1$, $w_{\rm cnf}=20$, $w_{\rm loc}=10$, and $w_{\rm sig}= 10$. We initialized the network's backbone and player's keypoints head using the pre-trained weights taken from PifPaf framework.

\paragraph{Quality measures.} 
This paper tackles a multi-objective task, hence, requires multiple quality measures for expressing the performance of individual sub-tasks. A detailed explanation of the quality measures is provided in the Supplementary Document. We define the ball Detection Quality (bDQ) as in the BallSeg framework~\cite{van2019real}, \ie the Area under the  Curve (AUC) associated with the Receiver Operating Characteristic (ROC) curve: the True Positive (TP) rate vs. the False Positive (FP) rate as a function of the detection score threshold.

We compute player Segmentation Quality (pSQ) and player Detection Quality (pDQ) identical to two components of the Panoptic Quality (PQ) measure introduced in~\cite{kirillov2019panoptic}, but restricted to only player (or person) class.

The pose Estimation Quality (pEQ) is measured based on the Object Keypoint Similarity (OKS) criteria defined in~\cite{Ronchi_2017_ICCV}. The components of pEQ measure are the variants of Average Precision (AP) and Average Recall (AR) thresholded at different OKS values.

\paragraph{Computational resources.} Time and memory performance were measured at inference on batches of one single $641\times 641$ pixels image, on machine configured with an NVidia V100 and a 32 cores Intel Xeon Gold 5217 running at 3GHz. Inference time is given in milliseconds~(ms) and GPU memory in megabytes~(MB).


\paragraph{Evaluation.}

We evaluate the performance of all three tasks on DeepSport's test set. Predicted and annotated human instances outside the basketball court are discarded during the computation of quality criteria.

Table \ref{tab:perf} shows that DeepSportLab only lacks 1\% in terms AP but improves by 2.5\% in terms of AR compared to \openpifpaf, which leads to an improvement of 0.5\% in terms of $F_1$ in total. In terms of player segmentation quality, DeepSportLab lacks by only 1.9\% and 1.3\% in terms of pSQ and pDQ, respectively, compared to Panoptic-DeepLab~\cite{cheng2020panoptic}. Since DeepSportLab outputs three different tasks at the same time, obtaining such a good segmentation quality reveals the excellence of the multi-task training.

\begin{figure*}[h!]
\centering
\begin{tabular}{c@{\hspace{1mm}}c@{\hspace{1mm}}c}
    \includegraphics[width=0.32\textwidth,trim={10.5cm 6.5cm 3cm 11.5cm},clip]{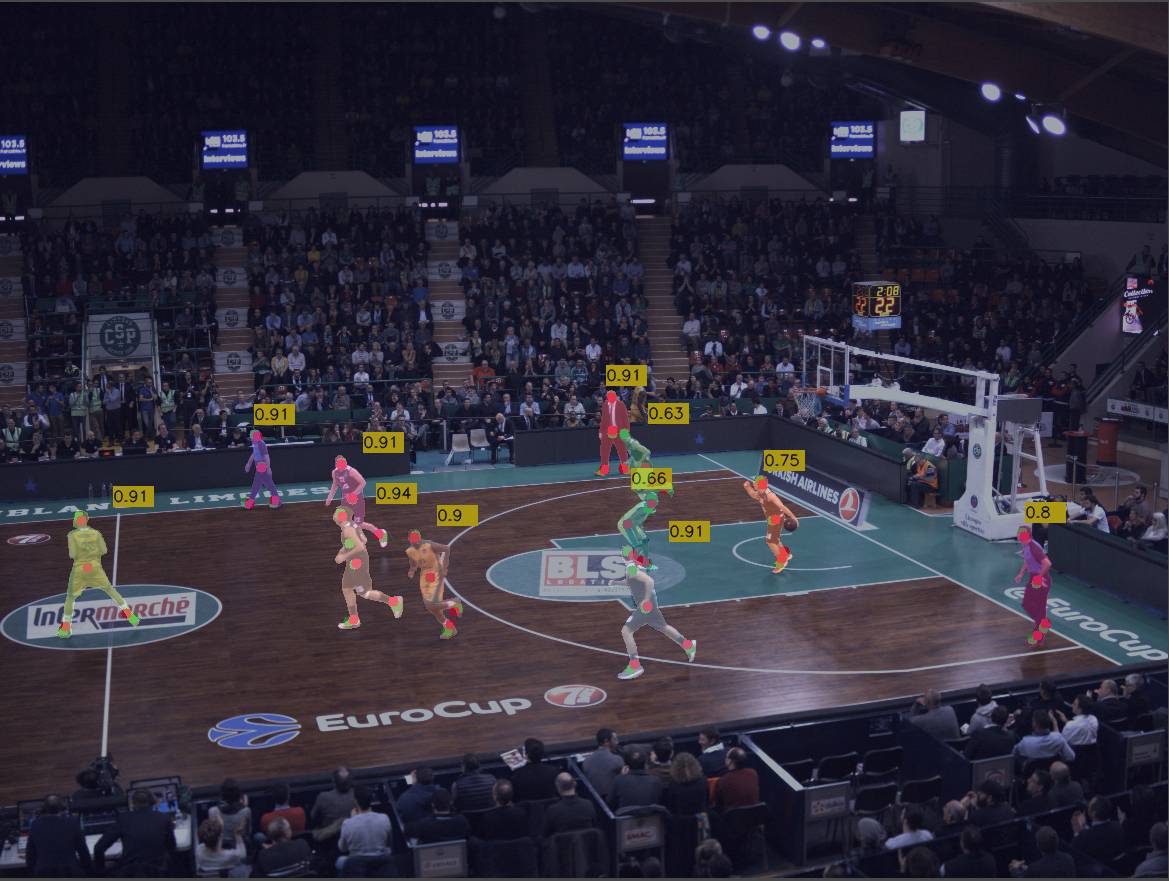}
    &
    \includegraphics[width=0.32\textwidth,trim={10.5cm 6.5cm 3cm 11.5cm},clip]{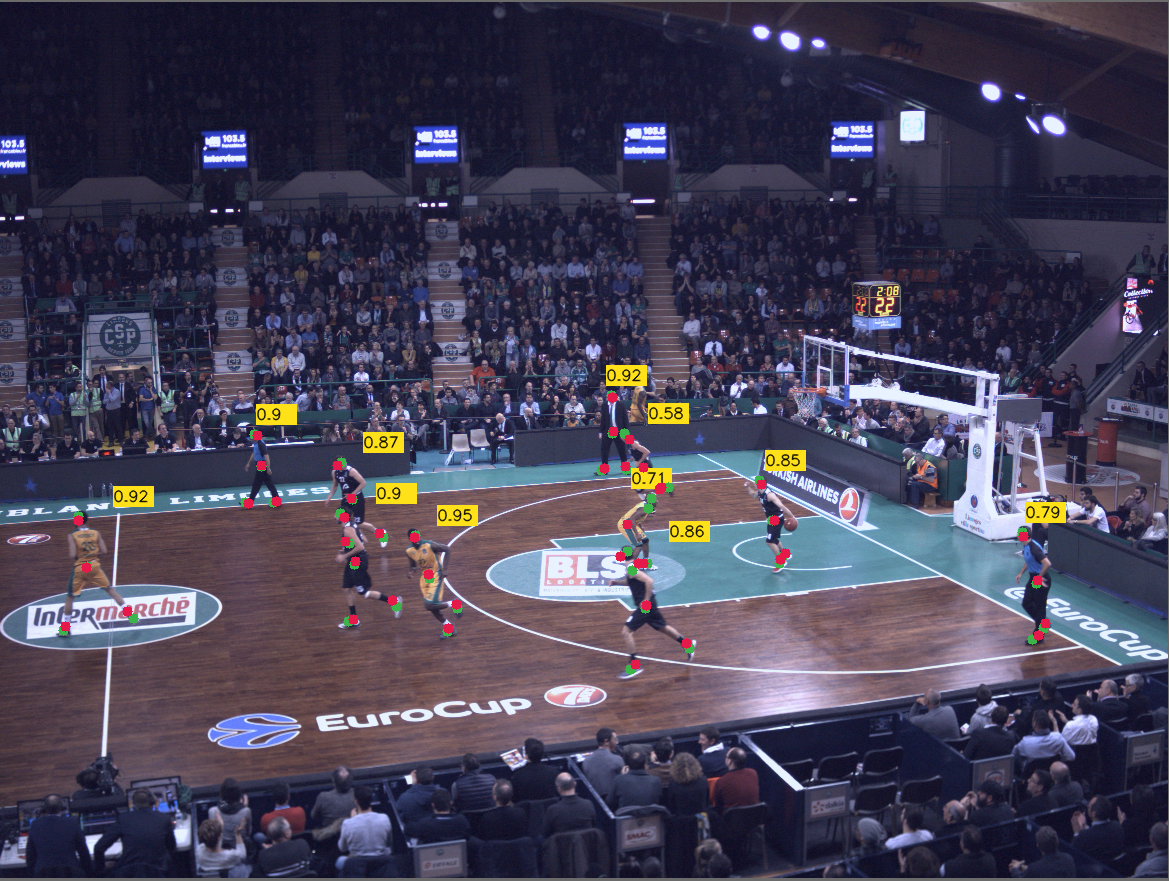}
    &
    \includegraphics[width=0.32\textwidth,trim={10.5cm 6.5cm 3cm 11.5cm},clip]{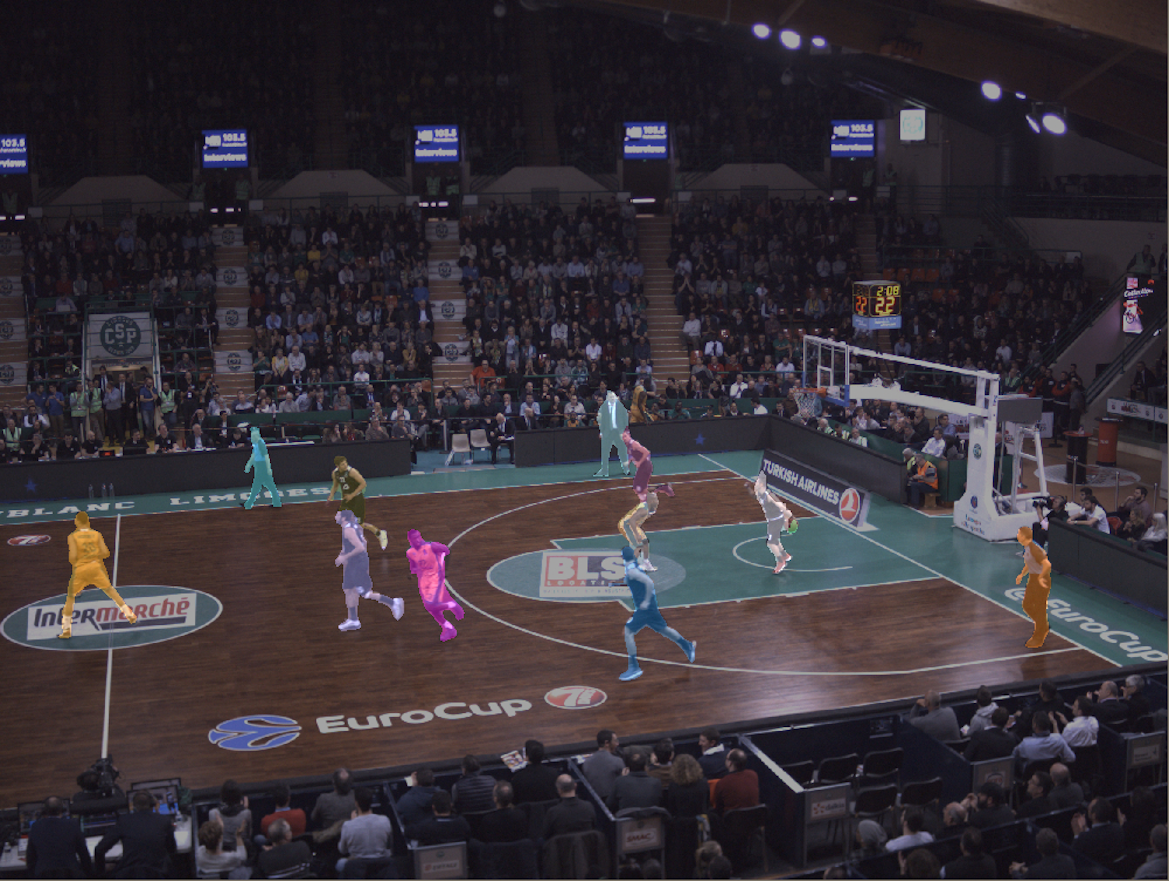}
    \\
    \footnotesize\textbf{(a) DeepSportLab}
    &
    \footnotesize\textbf{(b) PifPaf}
    &
    \footnotesize\textbf{(c) Panoptic-DeepLab}
\end{tabular}
\caption{\textbf{Pose recognition and mask segmentation samples.} Body parts are shown with red and green colors as prediction and ground truth, respectively. The numbers highlighted in yellow are OKS values for their corresponding prediction-annotation matching. In addition, colors are used to show the segmented player masks.}
\label{fig:pose-samples}
\end{figure*}

A visual analysis of our results compared to \openpifpaf\ and Panoptic-DeepLab~\cite{cheng2020panoptic} is presented in Fig.~\ref{fig:pose-samples}. It reveals that body part keypoints are generally assigned to their correct instance based on our computationally straightforward assignment strategy (see Sec.~\ref{sec:method-inference}). 


Regarding ball detection, our multi-task DeepSportLab framework is compared to the 
\gva{state-of-the-art BallSeg~\cite{van2019real}, which formulates the ball detection problem as a segmentation task. BallSeg uses an ICNet network~\cite{zhao2018icnet} trained on balls from the original DeepSport dataset~\cite{deepsport} (i.e. ball size ranges between 15 and 45 pixels). The performances  provided in Table~\ref{tab:perf} result from a training where images were scaled to keep humans with similar height compared to those found in the COCO dataset (i.e. ball size ranges between 7 and 18 pixels). At that scale, DeepSportLab significantly outperforms BallSeg}.

\gva{The computational comparison of the different models demonstrates the benefit of the unified framework. DeepSportLab can be deployed on devices with less than 2GB of memory, while the combination of its three counterparts would require almost 5GB. Inference time is only informative
due to the different levels of optimization of each implementation.
Actually, when combining different task-specific models, inference times do not add up when running computations in parallel. 
}

\section{Ablation studies}

\paragraph{DeepSportLab Decoder.}
Our proposed pose decoder is original in that it uses the segmentation mask to assign the keypoints to each instance. In Table~\ref{tab:decoder}, this strategy is applied on oracle data as well as on off-the-shelf predictions form \openpifpaf\ and Panoptic DeepLab~\cite{cheng2020panoptic}. 
The use of oracle masks instead of the intermediate predictions increases the $F_1$ Score by only 0.5\%. This shows that, \SAGHc{on the DeepSport dataset}, the instance segmentation is good enough to associate the player keypoints, meaning that the pose estimation task is only limited by the keypoints detection.
Using off-the-shelf models to produce the intermediate mask and the keypoints predictions does not improve the pose quality, revealing that our keypoint prediction is competitive. 
Additional investigations considering oracle data and error breakdown are provided in supplementary material.

\paragraph{Sport Specific Dataset.}
While containing a fair amount of balls, the COCO dataset is not rich enough to address the objective of DeepSportLab. This is illustrated in Table~\ref{tab:cocotestds}, where DeepSportLab was trained exclusively on COCO and evaluated on the DeepSport dataset~\cite{deepsport}. The ball detection quality (bDQ) value reported for COCO is significantly smaller than the one obtained when training on DeepSport. This demonstrates the need of having a task specific dataset to reach good performance when working on sport data.
\begin{table}
    \centering
    \scalebox{0.75}{
    \begin{tabular}{|l|c|cc|ccc|}
    	\hline
    	\multirow{2}{*}{Dataset used at training} & \multirow{2}{*}{$ {\rm bDQ}$} & \multirow{2}{*}{$ {\rm pSQ}$} & \multirow{2}{*}{$ {\rm pDQ}$} & \multicolumn{3}{c|}{$ {\rm pEQ}$} 
    	\\\hhline{|~~~~|---|}
    	  &	& & &  $ {\rm AP}$ & $ {\rm AR}$ & ${\rm F}_1$ \\
    	\Xhline{2pt}
    	\rowcolor{blue!10} COCO \& DeepSport & \gva{52.07} & 80.3 & 90.1 & 87.5 & 82.1 & 42.4 \\
    	\hline
    	 COCO & \gva{23.19} & \SAGHc{75.2} & \SAGHc{82.7} & \SAGHc{87.6} & \SAGHc{81.6} & \SAGHc{42.3}  \\
    	\Xhline{2pt}
    \end{tabular}
    }
\caption{\textbf{Sport-specific dataset study.} DeepSportLab evaluated on DeepSport dataset~\cite{deepsport} shows limited performances when trained exclusively on COCO dataset~\cite{lin2014microsoft}. This demonstrate the need for a sport specific dataset. Note: the ball detection quality (bDQ) reported when training only on COCO has been achieved by filtering out training images exempt of balls from the official COCO dataset. Without that filtering, bDQ evaluated on DeepSport dataset is far worse.}
    \label{tab:cocotestds}
\end{table}

\paragraph{DeepSportLab for individual tasks.}
Table~\ref{tab:individualtasks} evaluates the penalty or gain induced by the multi-task objective compared to an individual training of each task on the DeepSportLab backbone. The instance segmentation task benefits from the joined training, while the ball detection task does not. We understand this observation by the fact that the players mask are tightly coupled to their pose while the ball is not.

\begin{table*}[t]
    \centering
    \scalebox{0.75}{
    \begin{tabular}{|l|c|cc|ccc|cc|}
    	\hline
    	\multirow{2}{*}{Method}
    	&
    	\multirow{2}{*}{$ {\rm bDQ}$}
    	&
    	\multirow{2}{*}{$ {\rm pSQ}$}
    	&
    	\multirow{2}{*}{$ {\rm pDQ}$}
    	&
    	\multicolumn{3}{c|}{$ {\rm pEQ}$}
    	&
    	\multirow{2}{*}{$ {\rm ms}$}
    	&
    	\multirow{2}{*}{$ {\rm MB}$}
    	\\
    	\hhline{*{2}{|~|}*{2}{~}*{3}{|-}}
    	& &	& &  ${\rm AP}$ & ${\rm AR}$ & ${\rm F}_1$ && \\
    	\Xhline{2pt}
    	\rowcolor{blue!10}	DeepSportLab & \gva{52.07} & 80.3 & 90.1 & 87.5 & 82.1 & 42.4 & \SAGHc{$436 \pm 106$} & \SAGHc{1757} \\
    	\hline
    	DeepSportLab - Ball only & \gva{54.72} & -- & -- & -- & -- & -- & \SAGHc{$73 \pm 12$} &\gva{1688} \\\hline
    	DeepSportLab - Instances masks only & -- & \SAGHc{73.4} & \SAGHc{86.6} & -- & -- & -- & \SAGHc{$164 \pm 37$} & \SAGHc{1753}\\\hline
    	DeepSportLab - Poses (masks + keypoint) only & -- & \SAGHc{79.2} & \SAGHc{90.2} & \SAGHc{87.8} & \SAGHc{82.2} & \SAGHc{42.5} & \SAGHc{$496 \pm 107$} & \SAGHc{1755} \\
    	\Xhline{2pt}
    \end{tabular}
    }
\caption{\textbf{DeepSportLab on individual tasks study.} DeepSportLab was trained on the individual tasks and compared with the result of the combined training. Only the ball doesn't benefit from being trained jointly with another task.}
    \label{tab:individualtasks}
\end{table*}



\section{Conclusions}\label{sec:conclusion}

DeepSportLab is a framework handling pose estimation, instances segmentation and ball detection tasks, central to team sports analysis. 
Its architecture, based on a shared backbone, makes it more practical to deploy compared to existing off-the-shelf solutions, since it dramatically reduces the memory requirements, without affecting the prediction accuracy.
It proposes a new ball detection approach and a novel pose decoding algorithm based on the instance masks and showing interesting performances.
Two main lessons are drawn in terms of image interpretation problem formulation: \textit{(i)} adopting part intensity fields to locate the ball appears to be as effective than formulating this problem as a high-resolution image segmentation problem; and \textit{(ii)} assigning pose keypoints to their respective instances based on the spatial offsets predicted for instance segmentation sounds like a promising solution to reduce the decoding cost of PifPaf's pose recognition. 
In terms of perspectives, we observe that the bottleneck for more accurate player pose estimation lies in the prediction of keypoints, and not in their assignment to instances. Fundamental research is also desired to extend our multi-task framework to more generic datasets. 

\section*{Acknowledgments}\label{sec:acknowledgments}
Part of this work has been funded by the Belgian NSF, and by the Walloon region project DeepSport. 
C. De Vleeschouwer is a Research Director of the Fonds de la Recherche Scientifique - FNRS. S.A. Ghasemzadeh is funded by the FRIA/FNRS. M.~Istasse and G.~Van~Zandycke are funded by SportRadar.
Computational resources have been provided both by the supercomputing facilities of the Université catholique de Louvain (CISM/UCL) and the Consortium des Équipements de Calcul Intensif en Fédération Wallonie Bruxelles (CÉCI) funded by the Fond de la Recherche Scientifique de Belgique (F.R.S.-FNRS) under convention 2.5020.11 and by the Walloon Region, as well as by SportRadar private computing cluster.
The extended DeepSport dataset was provided by SportRadar.

\bibliographystyle{ieee_fullname}
\bibliography{deepsportlab}

\clearpage
\appendix

\section{Quality measures}\label{sec:sup:metric}
As mentioned in the main document, we tackle a multi-objective task, hence, requires multiple quality measures for expressing the performance of individual sub-tasks. The details of the quality measures used in our work is provided in the following.

\paragraph{Ball Detection Quality (bDQ).}
As in the BallSeg framework~\cite{van2019real}, given a threshold $\tau$ in the dynamic range of the confidence scores, a predicted ball keypoint is identified as a True Positive (TP) (or False Positive (FP)) detection , if its location lies inside (respectively, outside) the ground truth mask and its predicted confidence is greater than $\tau$: for the predicted ball $o^{\rm ball}$,
\begin{align}
&o^{\rm ball} \in {{\rm TP}^{\rm ball}}(\tau)\Leftrightarrow o^{\rm ball} \in \cl I^{{\rm ball}*},~\tilde{\bs S}({\rm ball}, o^{\rm ball}) \ge \tau\nonumber\\
&o^{\rm ball} \in {{\rm FP}^{\rm ball}}(\tau)\Leftrightarrow o^{\rm ball} \not\in \cl I^{{\rm ball}*},~\tilde{\bs S}({\rm ball}, o^{\rm ball}) \ge \tau
\end{align}
where $\cl I^{{\rm ball}*}$ denotes the semantic mask of the ball. By repeating this procedure for all images, we obtain the TP and FP sets associated with the full set of images. The TP rate (TPr) and FP rate (FPr) ratios are defined as

\begin{align*}\label{eq:metric-fp-rate}
{\rm TPr}(\tau) &\coloneqq \frac{|{\rm TP}^{\rm ball}(\tau)|}{|\{{\rm images ~with~annotated~ball}\}|},\\
{\rm FPr}(\tau) &\coloneqq \frac{|{\rm FP}^{\rm ball}(\tau)|}{|\{{\rm all~images}\}|}.
\end{align*}

The bDQ is then computed as the area under the ROC curve associated with the ${\rm TPr}$ and ${\rm FPr}$. 

\paragraph{Player Segmentation Quality (pSQ).}
Compared with the bDQ, a predicted player mask is identified as a TP mask (otherwise, an FP), if its IoU with one of the ground truth player masks is higher than the threshold of 0.5: 
\begin{equation}\label{eq:sq-tp}
(\cl I^{\rm ply}_i,\cl I^{{\rm ply}*}_j) \in {{\rm TP}^{{\rm ply}}} \Leftrightarrow \exists j,~{\rm IoU}(\cl I^{\rm ply}_i, \cl I^{{\rm ply}*}_j) \ge 0.5.
\end{equation}
The pSQ is then defined as the averaged IoU over the TP pairs:
\begin{equation}\label{eq:metric-psq}
{\rm pSQ} \coloneqq \frac{1}{|{\rm TP}^{\rm ply}|} \sum_{(u,v) \in {\rm TP}^{\rm ply}}{\rm IoU}(u,v).
\end{equation}

\paragraph{Player Detection Quality (pDQ).} Having identified the TP set as in \eqref{eq:sq-tp}, the pDQ is then defined as the $F_1$-score:
\begin{equation}\label{eq:metric-pdq}
{\rm pDQ} \coloneqq \frac{2|{\rm TP}^{\rm ply}|}{N_{\rm ply}+N^*_{\rm ply}}.
\end{equation}
Note that the pSQ~\eqref{eq:metric-psq} and pDQ~\eqref{eq:metric-pdq} criteria are the segmentation and recognition quality components of the Panoptic Quality (PQ) measure introduced in~\cite{kirillov2019panoptic}, \ie, concretely, the PQ for player segmentation reads ${\rm PQ} \coloneqq {\rm pSQ}\cdot {\rm pDQ}$.
\paragraph{Pose Estimation Quality (pEQ).} 
For pose estimation task we use the OKS criteria, \ie, for every pair of the predicted pose $\bs \Upsilon_i$ and ground truth pose $\bs \Upsilon^*_j$, it is defined as
\begin{equation}\label{eq:oks}
{\rm OKS}(\bs \Upsilon_i,\bs \Upsilon^*_j) \coloneqq {\rm mean}_{k} \exp(-\frac{\|o^k_i - o^{k*}_j\|_2^2}{2s^2_j\kappa^2_k}),
\end{equation}
where the mean is taken over the annotated body part keypoints, $s$ denotes the square root of the area of the bounding-box tightly containing all the body parts, and $\kappa_k$ is the per-keypoint-type scale constant controlling falloff.
The predicted skeletons are then sorted according to their confidence scores defined as the average over the body part confidence scores: from the pixel-wise confidence map in (Eq. 2 in the main document)
\begin{align}
\bs \Upsilon_i^{\rm conf} = \frac{1}{17} \sum_{k\in \cl K_{\rm part}}\tilde{\bs S}(k;o^{k}_i).\label{eq:pose-conf}
\end{align}

Next, the ordered predictions are assigned to the ground truths, with which they have the highest OKS value. Once the matching is complete, the set of TP skeletons $\rm TP^{skl}(\tau)$ with respect to the OKS threshold $\tau$ is determined. Concretely, for a fixed OKS threshold $\tau$ (ranging from 0.5 to 0.95), a pair of predicted and ground truth skeletons is identified as a TP, if their OKS is higher than $\tau$. The Precision (Pr) and Recall (Re) values are then computed as
\begin{equation}\label{eq:metric-P}
{\rm Pr}(\tau) \coloneqq \frac{|{\rm TP}^{\rm skl}(\tau)|}{N_{\rm ply}},~~~
{\rm Re}(\tau) \coloneqq \frac{|{\rm TP}^{\rm skl}(\tau)|}{N^*_{\rm ply}}.
\end{equation}
Finally, the Average Precision (AP) and Average Recall (AR) values read, respectively,
\begin{equation} \label{eq:metric-AP}
{\rm AP} \coloneqq {\rm mean}_\tau ~{\rm Pr}(\tau),~~~
{\rm AR} \coloneqq {\rm mean}_\tau ~{\rm Re}(\tau).
\end{equation}
\begin{rem}
The quality metrics above are defined per-image; however, in practice, we compute the bDQ, pSQ \eqref{eq:metric-psq}, pDQ~\eqref{eq:metric-pdq}, and AP and AR~\eqref{eq:metric-AP} over all the images in the validation or test sets.
\end{rem}

\begin{rem}
As required for computing pEQ, the values of $\kappa_k$ associated with the body parts are set according to the convention of DeepSport dataset, \ie, $\kappa_{\rm head} = 0.15$, $\kappa_{\rm hip} = 0.2$, and $\kappa_{\rm foot~1} = \kappa_{\rm foot~2}=0.2$.
\end{rem}

\begin{table*}[h!]
\centering
\scalebox{0.75}{
\begin{tabular}{|ccc|ccc|ccc|}
	\hline
	\small player &\small offset &	\small semantic	&\multirow{2}{*}{${\rm PQ}$}& \multirow{2}{*}{${\rm pSQ}$}&
	\multirow{2}{*}{${\rm pDQ}$}&
	\multicolumn{3}{c|}{${\rm pEQ}$}
	\\
	\hhline{|~~~|~~~|---|}
	\small centroid	&\small vectors &\small masks & & & & $ {\rm AP}$ & $ {\rm AR}$ & $ {\rm F}_1$ \\
	\Xhline{2pt}\rowcolor{blue!10}
	 --& --& -- & 72.3 & 80.3 & 90.1 & 87.5 & 82.1 & 42.4 \\
	 \hline
	 \checkmark& -- & -- & 71.4  &  79.7  & 89.6  & 87.2 & 82.7 & 42.5 \\
	 --& \checkmark & -- & 75.7 & 82.9  & 91.3    & 87  & 81.8  & 42.2 \\
	 --&--  & \checkmark &  86.7 & 94.5  & 91.7    & 87.2  & 82.5  & 42.4 \\
	 \hline
	 \checkmark & \checkmark & -- & 77 & 83 & 92.8 & 86.2 & 82.3  & 42.1 \\
	 \checkmark& -- & \checkmark & 87 & 94.5 & 92.1  & 87.7 & 83.7  & 42.8 \\
	 --& \checkmark & \checkmark & 93.5 & 98.3 & 95.1  & 88  & 82.7  & 42.6 \\
	 \hline
	\checkmark & \checkmark & \checkmark & 100 & 100  & 100  & 87.7  & 83.8 & 42.9 \\
	\Xhline{2pt}
\end{tabular}
}
\vspace{3mm}
\caption{\textbf{DeepSportLab Decoding with oracle data.} Note that ${\rm PQ} \coloneqq {\rm pSQ}\cdot {\rm pDQ}$.}
\label{tab:ablation}
\end{table*}

\begin{table*}[h!]
    \centering
    \scalebox{0.75}{
    \begin{tabular}{|l|ccc|ccc|}
    	\hline
    	\multirow{2}{*}{Method}               & \multirow{2}{*}{PQ} &
    	\multirow{2}{*}{pSQ} & \multirow{2}{*}{pDQ} & \multicolumn{3}{c|}{pEQ}  \\\hhline{~~~~---}
                                              &                      &                      & &   AP  &  AR  &  $F_1$  \\\Xhline{2pt}
    	\rowcolor{blue!10}	DeepSportLab      & 34.3 & 75.3                & 45.5                 & 43.7  & 44.2 & 22 \\\hline
    	DeepSportLab - oracle center & \SAGHc{52.1}   &  \SAGHc{78.1}           & \SAGHc{66.7}            & \SAGHc{57.1} & \SAGHc{60.7} & \SAGHc{29.4} 
    	\\\hline
    	DeepSportLab - oracle segmentation    &  \SAGHc{100}    &  \SAGHc{100}        & \SAGHc{100}            & \SAGHc{63.5} & \SAGHc{67.2} & \SAGHc{32.6}
    	\\\Xhline{2pt}
    	OpenPifPaf \cite{kreiss2021openpifpaf}    &  --   & --  & --     & \SAGHc{66.9} & \SAGHc{70.9} & \SAGHc{45.4}
    	\\\hline
    	Pan.-DeepLab~\cite{cheng2020panoptic} & \SAGHc{48.4}  & \SAGHc{78.6}       & \SAGHc{61.5}         & --    & --   & -- \\\Xhline{2pt}
    \end{tabular}
    }
    \vspace{3mm}
    \caption{\textbf{Comparison of different methods evaluated on COCO's validation set.} Three different cases are considered for DeepSportLab: (1) Decoding with network's outputs, (2) Decoding with oracle centers, and (3) Decoding with oracle segmentation masks.}
    \label{tab:player-coco}
\end{table*}

\section{Ablation studies}\label{sec:ablation}

\subsection{Decoding with Oracle Data} The importance of the accuracy of each output for the decoding process can be obtained by using the oracle data instead of their corresponding network outputs. This study is helpful to find out whether the error propagates from one block to the other. Table \ref{tab:ablation} compares the metrics when different permutations of oracle data were used on DeepSport dataset. The first message drawn from this study is that our player segmentation is good enough to associate the PIF keypoints because the increase of $F_1$ Score is only by 0.5\% when using all of the oracle data compared to when using none. Thus, for further improve the pose estimation task, the PIF keypoints should be trained better.

\subsection{Keypoints Error Breakdown} 
\begin{figure*}
    \centering
    \begin{minipage}{0.3\textwidth}
	\centering
	\scalebox{0.65}{\includegraphics{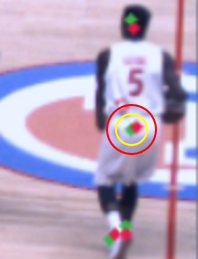}}
	\vspace{2mm}
	\caption*{\textbf{(a) Good}}
\end{minipage}
\begin{minipage}{0.3\textwidth}
	\centering
	\scalebox{0.65}{\includegraphics{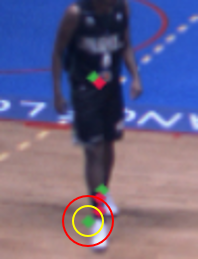}}
	\vspace{2mm}
	\caption*{\textbf{(b) Jitter}}
\end{minipage}
\begin{minipage}{0.3\textwidth}
	\centering
	\scalebox{0.65}{\includegraphics{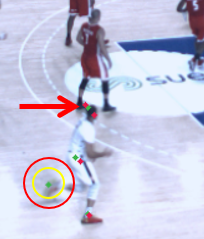}}
	\vspace{2mm}
	\caption*{\textbf{(c) Swap}}
\end{minipage}
\begin{minipage}{0.3\textwidth}
	\centering
	\scalebox{0.65}{\includegraphics{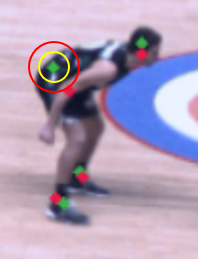}}
	\vspace{2mm}
	\caption*{\textbf{(d) Miss}}
\end{minipage}
\begin{minipage}{0.3\textwidth}
	\centering
	\scalebox{0.65}{\includegraphics{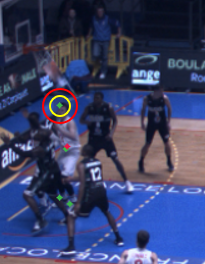}}
	\vspace{2mm}
	\caption*{\textbf{(e) FN KP}}
\end{minipage}
\vspace{2mm}
    \caption{\textbf{Error samples.} Green and red dots show the annotated and predicted keypoints, respectively. Yellow and Red circles resemble the borders from which the KS will be less than 0.85 and 0.5, respectively. In (c), the red arrow points toward the wrongly predicted foot where the swap occurs.}
    \label{fig:error}
\end{figure*}

\begin{figure*}
\centering
    \begin{minipage}{\linewidth}
        \centering
        \begin{minipage}{0.29\linewidth}
        \centering
        \scalebox{0.35}{\includegraphics{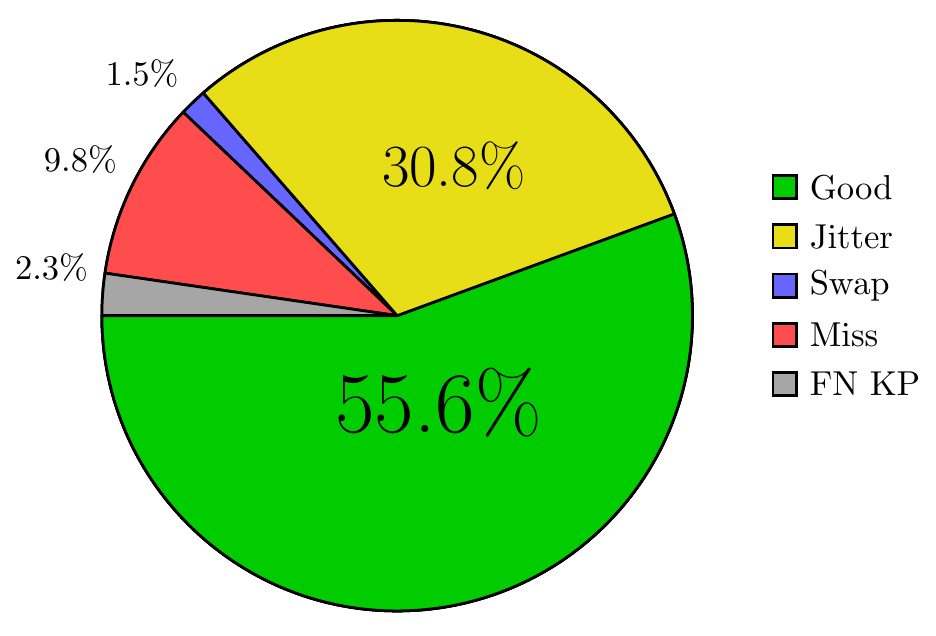}}
        \end{minipage}
        \begin{minipage}{0.69\linewidth}
        \centering
        \scalebox{0.3}{\includegraphics{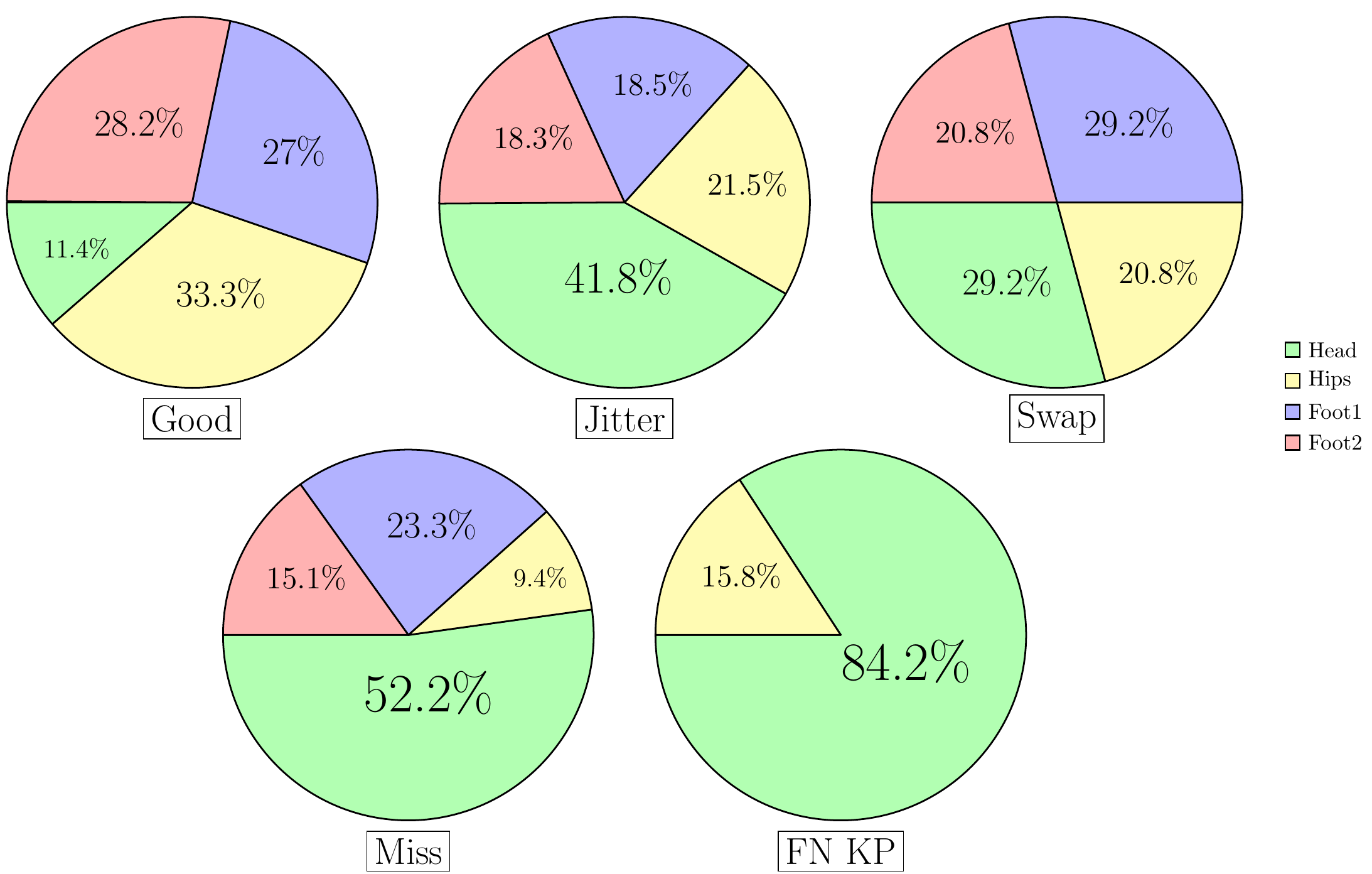}}
        \end{minipage}
    \vspace{4mm}
    \caption*{\textbf{(a) DeepSportLab}}
    \end{minipage}
    \begin{minipage}{\linewidth}
        \centering
        \begin{minipage}{0.29\linewidth}
        \centering
        \scalebox{0.35}{\includegraphics{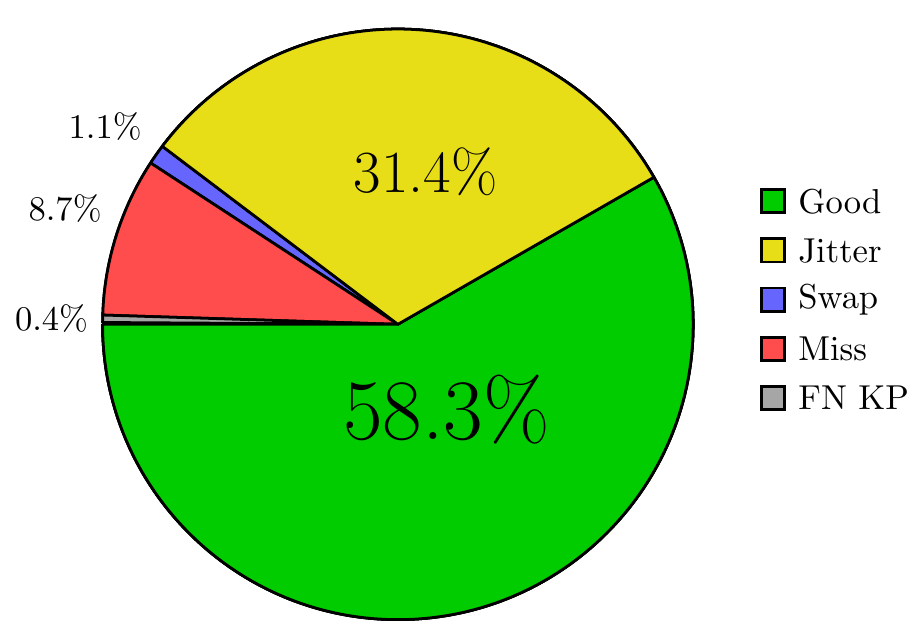}}
        \end{minipage}
        \begin{minipage}{0.69\linewidth}
        \centering
        \scalebox{0.3}{\includegraphics{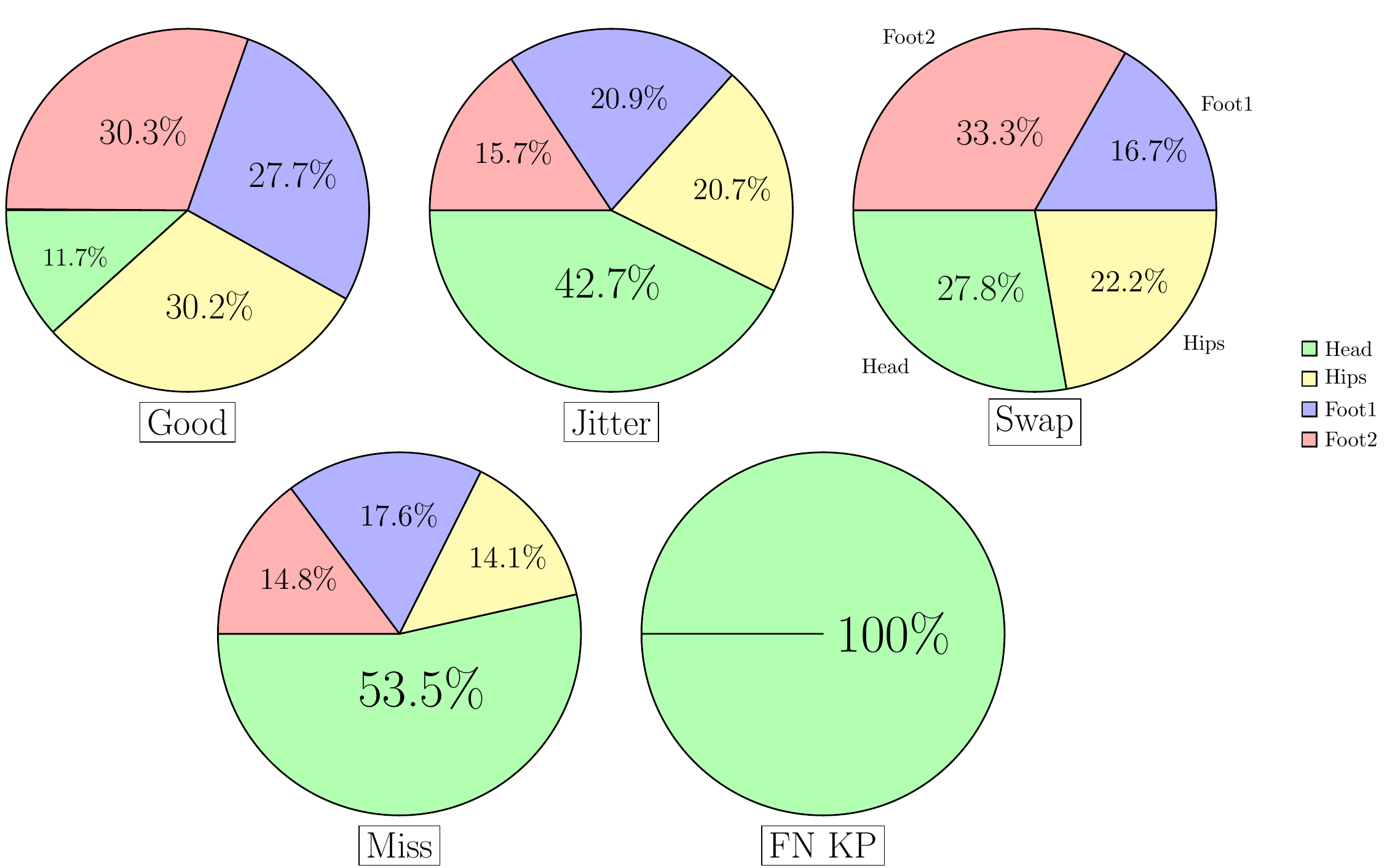}}
        \end{minipage}
    \vspace{4mm}
    \caption*{\textbf{(b) PifPaf}}
    \end{minipage}
    \vspace{2mm}
\caption{\textbf{Error breakdown.} Pie charts on the left show the distribution of keypoints in 5 categories based on their KS. Pie charts on the right show the distribution of each type of error based on the keypoints type.}
\label{fig:pie}
\end{figure*}

Further improvements can be achieved once we know the source of error. Ronchini and Perona~\cite{Ronchi_2017_ICCV} break the estimated body keypoints in 5 different categories based on their calculated KS, \ie keypoint similarity between the  keypoint $o$ of a detection $\bs \Upsilon$ and $o^*$ of an annotation $\bs \Upsilon^*$. KS is calculated using \eqref{eq:oks} without the mean over all keypoints. If KS of $o$ and $o^*$ is higher than 0.85, this prediction is considered Good. Jitter happens when KS drops between 0.5 and 0.85. In case that KS is less than 0.5, $o$ can be either a Miss, Swap, or Inversion. In our case, since we switch the right and left feet in case of wrong detection, the Inversion will never occur. This is because foot 1 and foot 2 labels are interchangeable in DeepSport dataset (See Section 4 in the main document). Next, Swap happens when $o$ is wrongly associated to another skeleton. Miss happens when $o$ is predicted, but not in the right location, and it was not a Swap. Finally, FN KP happens when the keypoint is not detected at all. Fig. \ref{fig:error} shows the examples for each of these categories. Fig. \ref{fig:pie} depicts the error breakdown based on the error category and type of keypoint.

\subsection{Evaluation on COCO dataset} 
As stated in the main text, our main contribution is to come up with a multi-task framework specific to sports scenes. 
However, as an ablation study, the model was also evaluated on COCO dataset~\cite{lin2014microsoft} which is very much diverse in terms of both the scenery and the size of people in images.
Table~\ref{tab:player-coco} shows metrics evaluated on COCO's validation set. Note that in this experiment, only the keypoints visible in the image are considered for the evaluation.

DeepSportLab decoder is studied in three different cases:
(1) When using the network outputs,
(2) When using oracle centroid of humans, and
(3) When using the full human mask oracle.
In the first case, due to the diversity of people size in COCO images, the segmentation task falls short in terms of pDQ which leads to error in pEQ. When adding the oracle center, the PQ increases significantly, suggesting that the center needs more training. When using the oracle masks, (\ie{} PQ = 100), pEQ increases by 45.3\% and 34.2\% in terms of AP and AR, respectively. This shows the importance of the segmentation masks on big and challenging datasets such as COCO. It is worth mentioning that training in this case needs a lot of hyper-parameter tuning and optimization. Our computational resources certainly did not allow to fully explore the parameter space. Obtaining more competitive results on COCO dataset is seen as a future work for this framework.

\end{document}